\documentclass[conference]{IEEEtran}
\IEEEoverridecommandlockouts
\usepackage{cite}
\usepackage{amsmath,amssymb,amsfonts}
\usepackage{algorithmic}
\usepackage{graphicx}
\usepackage{textcomp}
\usepackage{float}
\usepackage{xcolor}
\usepackage{booktabs}
\usepackage{multirow}
\usepackage[lined,boxed,linesnumbered, ruled]{algorithm2e}
\usepackage{subfigure}
\usepackage[switch]{lineno}

\def\BibTeX{{\rm B\kern-.05em{\sc i\kern-.025em b}\kern-.08em
    T\kern-.1667em\lower.7ex\hbox{E}\kern-.125emX}}
\begin{document}

\title{ A GAN-based data poisoning framework against anomaly detection in vertical federated learning\\
}
\author{
    \IEEEauthorblockN{Xiaolin Chen, Daoguang Zan, Wei Li, Bei Guan$^{*}$, Yongji Wang}
    \IEEEauthorblockA{Institute of Software, Chinese Academy of Sciences, Beijing, China}
    \IEEEauthorblockA{University of Chinese Academy of Sciences, Beijing, China}
    \IEEEauthorblockA{\{chenxiaolin2019, daoguang, guanbei\}@iscas.ac.cn, liwei224@mails.ucas.ac.cn, yj\_wang\_iscas@126.com}
}
\maketitle
\begin{abstract}
In vertical federated learning (VFL), commercial entities collaboratively train a model while preserving data privacy. However, a malicious participant's poisoning attack may degrade the performance of this collaborative model. The main challenge in achieving the poisoning attack is the absence of access to the server-side top model, leaving the malicious participant without a clear target model. To address this challenge, we introduce an innovative end-to-end poisoning framework P-GAN. Specifically, the malicious participant initially employs semi-supervised learning to train a surrogate target model. Subsequently, this participant employs a GAN-based method to produce adversarial perturbations to degrade the surrogate target model's performance. Finally, the generator is obtained and tailored for VFL poisoning. Besides, we develop an anomaly detection algorithm based on a deep auto-encoder (DAE), offering a robust defense mechanism to VFL scenarios. Through extensive experiments, we evaluate the efficacy of P-GAN and DAE, and further analyze the factors that influence their performance.
\end{abstract}

\section{Introduction}
Data plays an essential role in advancing machine learning. Constrained by global concerns about data security and implementation of related regulations~\cite{EU}, institutions are not permitted to share data for modeling directly. This highlights the necessity of exploring methods for multi-party collaborative modeling without disclosing their private data. 
Federated learning (FL)~\cite{mcmahan2017communication} as a promising solution, is proposed and has drawn considerable interest from both industry~\cite{niknam2020federated} and academia\cite{konevcny2016federated,kairouz2021advances}.

According to the classification proposed by Yang~\cite{yang2019federated}, FL can be divided into horizontal federated learning (HFL) and vertical federated learning (VFL). HFL encompasses scenarios where participants have distinct data samples but share a common feature space. In contrast, VFL is designed for situations where participants possess separate feature spaces but share a common sample space. This difference makes VFL well-suited for collaborative modeling among entities with different business attributes. In addition, VFL has been applied in many domains, such as financial risk management~\cite{cheng2021secureboost} and advertising~\cite{wei2021autoheri}.

With the growing adoption of vertical federated learning, researchers have become increasingly concerned about its security. DLG~\cite{zhu2019deep} was pioneering in revealing the risk of gradient leakage in deep learning. This work suggests that shared gradients might inadvertently expose intricate data features, allowing the malicious participant to reconstruct the original data. Subsequently, various inference attacks for both training and inference stages have been proposed. During training, RSA~\cite{weng2020privacy} and LIA~\cite{fu2022label} launch inference attacks on ensemble trees and neural networks. In the inference phase, GRNA~\cite{luo2021feature} and FIA~\cite{yang2023practical} employ generative models and zeroth-order gradient estimation methods to achieve feature inference. These methods pose threats to the privacy of VFl. Moreover, a growing concern is data tampering, such as the poisoning attacks discussed in this study.

Within VFL, some poisoning attacks arise when an adversary injects the tampered data into the training dataset. The tampered data disrupts the server-side top model during the training phase, resulting in decreased performance. However, the architecture of VFL makes the top model elusive to the malicious participant, i.e., the malicious participant lacks a target model for formulating the poisoning attack. LRA~\cite{liu2021batch} and VILLAIN~\cite{bai2023villain} poison the local date using fixed triggers and label replacement without a target model. However, these perturbation strategies do not fully capitalize on image representations. To address this limitation, we present a poisoning attack strategy based on the model completion technique. Our contributions can be listed as follows:
\begin{itemize}
\item We introduce P-GAN, an end-to-end poisoning attack mechanism for VFL. Specifically, by semi-supervised learning, we can obtain a surrogate target model in our approach during training. Subsequently, the poisoning model based on a generative adversarial network (GAN) generates perturbations by degrading the performance of the surrogate target model. These perturbations are added on the local features, serving as poisoned inputs. As VFL training, it degrades the performance of the top model.
\item To defend against poisoning attacks, we develop a server-side anomaly detection algorithm based on a deep auto-encoder (DAE). This algorithm filters out outliers in embedding vectors with reconstruction errors.
\item Through extensive experiments on MNIST, CIFAR-10, and CIFAR-100, we evaluate the performance of P-GAN and DAE. Our results indicate that P-GAN outperforms state-of-the-art approaches, with DAE effectively defending against poisoning attacks. Furthermore, we analyze the potential factors that influence their performance.
\end{itemize}


\section{Background and Related Work}
In this section, we describe the training process and the poisoning procedure in VFL, followed by the related works on poisoning attacks.
\subsection{The Data Poisoning Attack in VFL}
\begin{figure}[htbp]
\centerline{\includegraphics[width=0.85\linewidth]{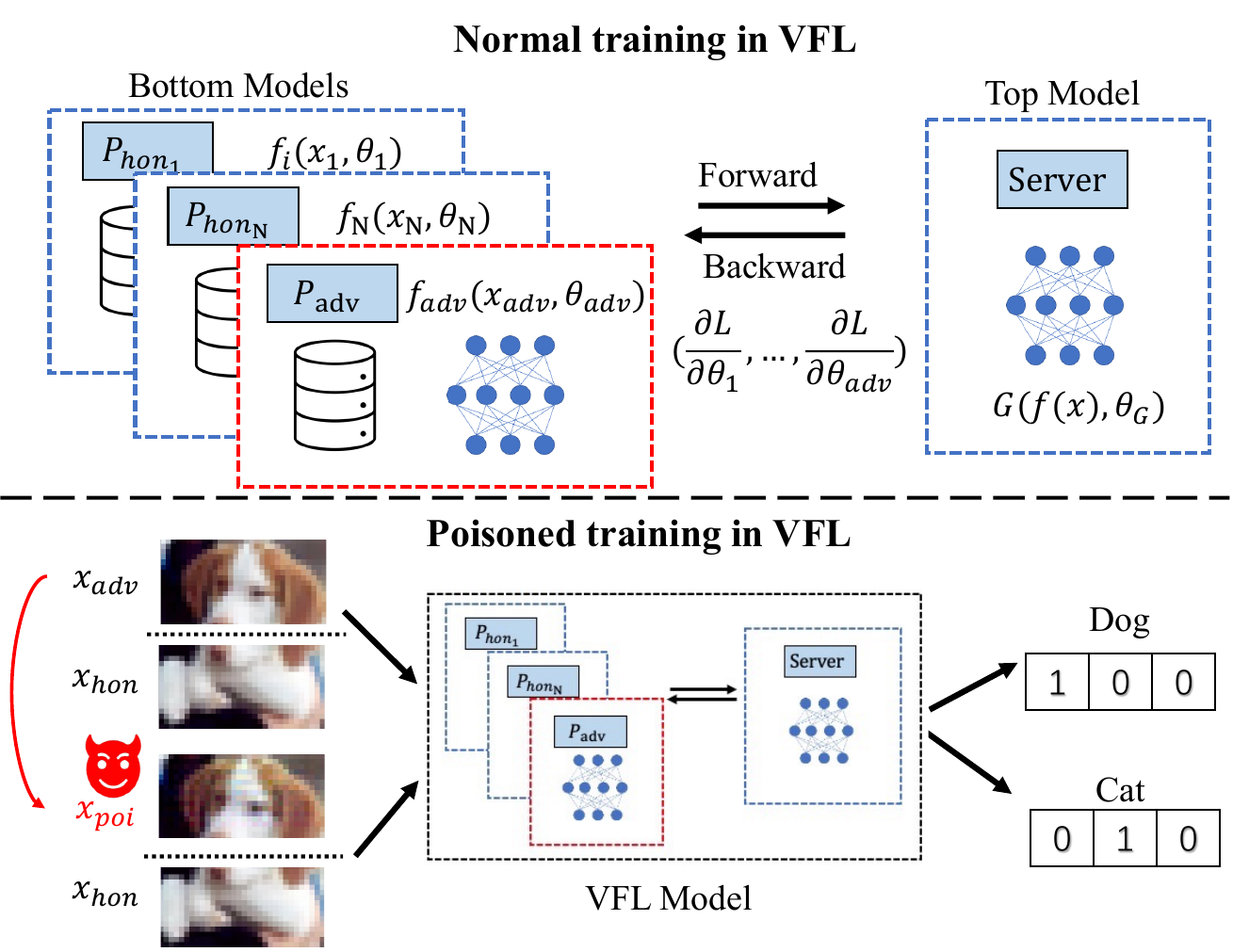}}
\caption{Vertical federated learning training process. The malicious participant (red) can attack the training process through data poisoning, causing the VFL model to classify ``dog" as ``cat".}
\label{VFL}
\end{figure}
In VFL, multiple participants with distinct features about the same user set collaboratively train machine learning models without disclosing their local data or local bottom models. Typically, the server retains the local model and labels, while each participant holds local feature subsets and the bottom model. 
As shown in Figure \ref{VFL}, during the normal training process, participants calculate their local model's embedding vectors $f_i(x_i)$ and submit them to a centralized server. The server concatenates these vectors, completes the forward propagation, and calculates the loss with labels. Subsequently, the server performs backward propagation and sends gradients to each local participant for model updates. The training process continues until the model converges. Figure \ref{VFL} illustrates the poisoned training process in VFL. The malicious participant injects poison into local data, degrading the performance of the server-side top model during training.
\subsection{Releated Work}
Recent studies have explored poisoning attacks in VFL. This study~\cite{yu2023backdoor} proposes backdoor attacks for both server and local participants. However, they assume poisoned data is already present during training without exploring its generation specifically for VFL. LRA~\cite{liu2021batch}, a black-box poisoning method, uses a label-replacement technique to create poisoned data. However, this label-replacement approach is easily defended since the server can detect inconsistencies between the embedding vectors and labels. Bai et al.~\cite{bai2023villain} introduces a data poisoning technique that designs a trigger backdoor by adding a fixed pattern on embedding vectors. The patterns are designed as follows:
\begin{equation}
\begin{aligned}
f_{adv}(x_{poi})=f_{adv}(x_{adv})\oplus\epsilon, \quad
\epsilon=\mathcal{M}\otimes(\beta*\Delta)
\end{aligned}
\end{equation}
where $\Delta$ represents a pattern of two positive values followed by two negative values, $\beta$ is the parameter that controls the trigger's magnitude, and $\mathcal{M}$ denotes the trigger mask.
\section{Poisoning attack model}

\begin{figure*}[ht]
\centerline{\includegraphics[width=0.725\linewidth]{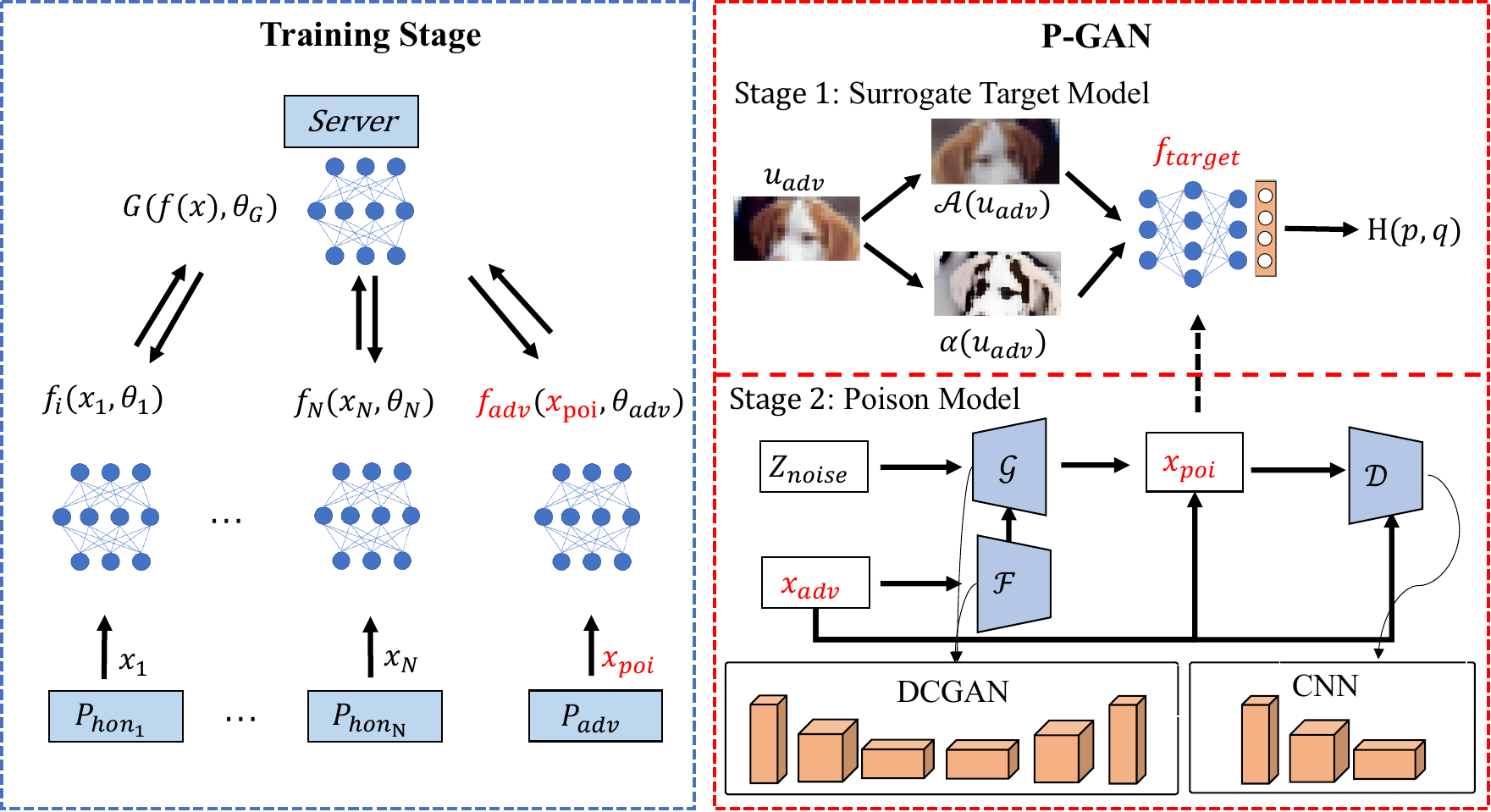}}
\caption{The framework of data poisoning attack in VFL (P-GAN). The malicious participant (red) can manipulate the model's training process by generating poison samples with the two-stage P-GAN algorithm.}
\label{fig1}
\end{figure*}
In this section, we first introduce the threat model for the poisoning attack in VFL, followed by the P-GAN framework designed for data poisoning.
\subsection{Threat model}
We assume that N+1 local participants and a server jointly train a machine learning model in VFL. Each local participant contribute partial features $\mathbf{x}_i$ for the machine learning model and we have $\mathbf{x}=(\mathbf{x}_1,...,\mathbf{x}_N,\mathbf{x}_{adv})$, the server holds the labels $y$ of dataset. Local participants include honest participants $P_{hon_i}(i=1,...,N)$ and a malicious participant $P_{adv}$. These local participants upload the outputs of bottom models $f_{i}(\mathbf{x}_i)$ to the server. The server predicts with the top model $G(f(\mathbf{x}))=G(f_{hon}(\mathbf{x}_{hon}),f_{adv}(\mathbf{x}_{adv}))=G(f_1(\mathbf{x_1}),...,f_N(\mathbf{x}_N),f_{adv}(\mathbf{x}_{adv}))$ and updates both the top model and bottom models. The goal of the malicious participant $P_{adv}$ is to degrade the performance of the top model by injecting noise $\epsilon$ into its local features $x_{adv}$. Besides, the injected perturbations $r$ should be as small as possible to avoid possible anomaly detection. Mathematically, given a threshold $\alpha$, the object can be expressed as follows:
\begin{equation}
\min_{\epsilon} \| r \|\qquad \text{s.t.}\ M_{x_{adv}}-\widetilde{M}_{x_{adv}+\epsilon} > \alpha 
\end{equation}
where $M_{x_{adv}}$ and $\widetilde{M}_{x_{adv}+\epsilon}$ define the performance metrics of the clean and poisoned top models. We define the knowledge and the capability of the adversary under a black-box assumption as follows:

\textbf{Adversary's knowledge.} During training, the adversary has access to its local features $\mathbf{x}_{adv}$ and the corresponding bottom model $f_{adv}$ but lacks knowledge of other participants' features $\mathbf{x}_{i}(i=1,...,N)$ and the server's labels $\mathbf{y}$.
Additionally, they can access training parameters, including sample indices, epochs, and gradient updates related to the bottom model.

\textbf{Adversary's capability.} During training, the adversary can train a local bottom model $f_{adv}$ based on $\mathbf{x}_{adv}$, which is a part of the VFL model. The adversary can upload the embedding vectors to the server and achieve the attack through multiple epochs. While it has the capability to tamper local features with the bottom model, they do not have access to the server-side top model. Besides, the adversary might infer a small amount of labels based on local data and gradients~\cite{fu2022label,bai2023villain}.

\subsection{P-GAN model}\label{sec:P-GAN}
Based on the adversary's knowledge, the malicious participant lacks access to the top model, preventing its direct use as a target model. Thus, we train a surrogate target model by model completion. With a limited set of labeled samples, the malicious participant constructs a local target model for the poisoning attack through Fixmatch~\cite{sohn2020fixmatch}. Subsequently, we train the poisoning model with the surrogate target model. Figure \ref{fig1} shows the framework of P-GAN.

\textbf{Stage 1: Surrogate target model.} 
During training, the malicious participant continuously updates the local bottom model $f_{adv}$. We can append a classification layer on the bottom model, using it as the initialization for the surrogate target model. We then employ the Fixmatch method, fine-tuning the surrogate target model with local features and a small set of labels. Fixmatch combines consistency regularization and pseudo-labeling, makes strong and weak augmentations on unlabeled data, and updates the model based on the consistency loss of classification results. Thus, we obtain the surrogate target model, which has similar predictions to the top model. Considering the transferability of adversarial examples~\cite{liu2016delving}, we can use the surrogate model as the target model $f_{target}$ for our poisoning attack.

\textbf{Stage 2: Poison model.} In P-GAN, we design a poisoning approach poison method based on a deep convolutional generative adversarial network (DCGAN) to minimally perturb data and induce misclassification of the surrogate target model. Specifically, the malicious participant generates the perturbations based on the embedding extract from the local features, following which perturbations are added on local features to produce poisoned data $x_{poi}=x_{adv}+\mathcal{G}(x_{adv}, Z_{noise})$. After this, upon uploading the poisoned data to the target model, the adversary loss $\mathcal{L}_{adv}$ is computed, while the discriminator categorizes poisoned and clean data. Then, both the discriminator $\mathcal{D}$ and generator $\mathcal{G}$ collaboratively adjust their parameters based on the total loss $\mathcal{L}_{total}$ until convergence. Ultimately, we construct a poisoning attack model locally for the malicious participant, enabling poisoning attacks during VFL training.

The overall procedure of P-GAN algorithm is presented in Algorithm \ref{P-GAN}. Where $\mathcal{A}(u_{adv}^b)$ and $\alpha(u_{adv}^b)$ represent strong and weak augmentations on unlabeled data. $p_{m}(x)$ denotes the predicted distribution of model $f_{target}$ on features $x$.

\SetNlSty{textbf}{}{:}
\IncMargin{0.1em}
\begin{algorithm}[htbp]\label{P-GAN}
\caption{P-GAN.}
\SetKwInOut{Input}{Input}
\SetKwInOut{Output}{Output}
\Input{Labeled examples $(x_{adv}^b,p^b), b=1,...,B$, unlabeled examples $(u_{adv}^b),b=1,...,\mu B$\ 
loss weight $\lambda_{u}$, $\lambda_{GAN}$, $\lambda_{r}$,
unlabeled data radio $\mu$ }
\Output{Trained generator $\mathcal{G}$ and discriminator $\mathcal{D}$}
\textbf{Stage 1: Surrogate target model}\\
Initialize the target model $f_{target}$ with the bottom model$f_{adv}$\\
Calculate the Cross-Entropy loss of labeled data
$\mathcal{L}_s=\frac{1}{B}\sum\limits_{b=1}^{B}H(p_{adv}^b, p_{m}(\alpha(x_{adv}^b)))$;\\
Predict class distribution for weakly or strongly augmented unlabeled data;\\
\For{$b=1,...,\mu B$}{
 $\widetilde{u}_{adv}^b=p_m(y|\mathcal{A}(u_{adv}^b))$;\\
 $q_{adv}^b=p_m(y|\alpha(u_{adv}^b))$;\\
}
Calculate the loss of unlabeled data 
$\mathcal{L}_u=\frac{1}{\mu B}\sum\limits_{b=1}^{\mu B} \mathbb I\{max(q_{adv}^b)>\tau\}H(arg\ max(q_{adv}^b),\widetilde{u}_{adv}^b)$; \\
Update model $f_{target}$ with loss $\mathcal{L}_s+\lambda_{u}\mathcal{L}_u$ and repeat;\\
~\\
\textbf{Stage 2: Poison attack model}\\
Load pre-train model $f_{adv}$ without classify layer;\\
Generate perturbation and poisoned data $x_{poi}=x_{adv}+\mathcal{G}(x_{adv},Z_{noise})$;\\
Calculate the loss of poisoned data $\mathcal{L}_{adv}=H(f_{target}(x_{poi}),target)$;\\
Discriminator $\mathcal{D}$ categorize $x_{adv}$ and $x_{poi}$ $\mathcal{L}_{GAN}=\mathbb E_x[log\mathcal{D}(x)]+\mathbb E_x[log(1-\mathcal{D}(x+\mathcal{G}(x)))]$;\\
Update the generator $\mathcal{G}$ and the discriminator $\mathcal{D}$ with total loss and repeat
$\mathcal{L}_{total}=\mathcal{L}_{adv}+\lambda_{GAN}\mathcal{L}_{GAN}+\lambda_r\mathbb E_x[\mathcal{G}(x)]$;
\end{algorithm}

\section{Defense Model}
In this section, we introduce an anomaly detection model against data poisoning attacks in VFL.

\begin{figure}[htbp]
\centerline{\includegraphics[width=0.85\linewidth]{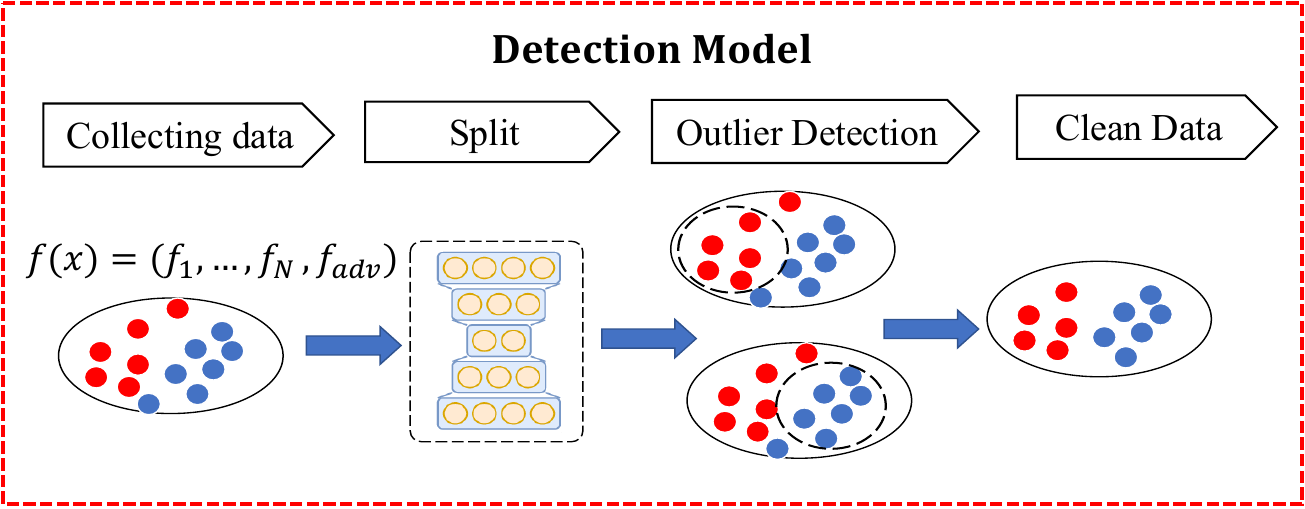}}
\caption{Flowchart of the proposed defence algorithm in VFL. The server employs the DAE method to filter out outliers for each class of embedding vectors.}
\label{fig2}
\end{figure}
In server-side anomaly detection, the server receives embedding vectors $(f_1(x_1),...,f_N(x_N),f_{adv}(x_{adv}))$, which may contain poisoned data. The server can identify anomalies in the vectors with its labels and reconstruction errors. Figure \ref{fig2} illustrates the overall procedure of anomaly detection in VFL. Initially, the server collects and concatenates embedding vectors uploaded from local participants. Subsequently, the server splits these vectors based on local ground-truth labels, ensuring data with the same labels cluster together. Considering the possibility of poisoned data in the training set, we opt for the DAE for outlier detection, which shows robustness against varying proportions of poisoned data~\cite{bovenzi2022data}. The encoder captures features of embedding vectors, while the decoder is responsible for its reconstruction. Anomalies are identified based on the reconstruction error of vectors, measured by root mean squared error (RMSE). Data is marked anomalous if the error exceeds a label-specific threshold. Finally, the vectors with a reconstruction error below the threshold are utilized as the clean training data on the server.

\section{Experimental Evaluation}
In this section, we first describe the experimental datasets and setup. We then evaluate the performance of P-GAN and DAE, followed by analyzing influential factors, such as the number of adversary's features, the quantity of known labeled data, and the proportion of poisoned samples.
\subsection{Datesets}
Our experiments are conducted on MNIST, CIFAR-10, and CIFAR-100. The summary of these datasets is as follows:
\begin{itemize}
    \item \textbf{MNIST:} It is a widely used dataset of handwritten digital images for image classification tasks, containing 60,000 training samples and 10,000 test samples, each of which is a gray-scale image of 28 × 28 pixels;
    \item  \textbf{CIFAR-10:} It is commonly used for image classification tasks and is a dataset of color images of 10 different classes, each of which is 32 × 32 pixels, containing a total of 50,000 training samples and 10,000 test samples;
    \item  \textbf{CIFAR-100:} It is also a popular dataset for image classification, encompassing 60,000 training images and 10,000 test images. Each image is a 32x32 pixels picture distributed across 100 classes.
\end{itemize}
\subsection{Experimental Setup}
In our experiments, we establish a VFL environment with two local participants and a central server. Both participants manage local model computations and data uploads. One acts as a malicious participant, aiming to degrade the performance of the federated model with its poisoned features. Data is partitioned such that each participant retained samples with half of the features, ensuring an equitable environment for performance evaluation.

Table \ref{tab:0} shows an overview of the datasets and their associated model architectures. Both the bottom and surrogate target models employ either a 3-layer fully connected network or ResNet structures. While the top model is a fully connected network, the defense model follows a standard DAE structure. The attack model, based on DCGAN, comprises three convolutional layers and three deconvolutional layers.
\begin{table}[htbp]
\renewcommand{\arraystretch}{1.1}
\caption{Model architectures.}
\vskip -0.1in
\label{tab:0}
\centering
\begin{tabular}{ccccc}
\toprule
Dataset & \begin{tabular}[c]{@{}c@{}}Bottom \\Model\end{tabular} & \begin{tabular}[c]{@{}c@{}}Top \\ Model\end{tabular} & \begin{tabular}[c]{@{}c@{}}Defense \\ Model\end{tabular} & \begin{tabular}[c]{@{}c@{}}Poisoning \\ Attack Model\end{tabular} \\ \hline
MNIST & FCNN-3 & FCNN-3 & DAE-7 & DCGAN-6 \\
CIFAR-10 & ResNet-18 & FCNN-3 & DAE-7 & DCGAN-6 \\
CIFAR-100 & ResNet-18 & FCNN-4 & DAE-7 & DCGAN-6 \\ 
\bottomrule
\end{tabular}
\end{table}

In our experiments, we select LRA and VILLAIN as the baseline algorithms. Additionally, we evaluate the effectiveness of DAE in detecting poisoning attacks. In our study, we use $F_1$ score and accuracy to evaluate the effectiveness of the attacks. A lower $F_1$ score or accuracy for a poisoned model indicates a more successful poisoning effect.

\subsection{Experimental Result}
We use the trained bottom model at the 5-th epoch as the initialization for the local target model. We then append a classification layer on the target model and train the model with limited labeled data by FixMatch. Comparing the true labels with predicted labels, we calculate the accuracy of the surrogate target model. The classification performance of the target model across various amounts of labeled samples is presented in the table \ref{tab:1} below. Despite having a limited number of labeled samples, the surrogate target model still has a commendable classification performance. When the quantity of labeled samples exceeds 160, the model's accuracy surpasses 70\% on the training dataset and 60\% on the test dataset. Considering the total sample size ranges from 50,000 to 60,000, 160 labeled samples represent a small and acceptable fraction. The attacker could infer a small subset~\cite{fu2022label} of labels with local features. Thus, model competition is both practical and acceptable.
\begin{table}[htbp]
\renewcommand{\arraystretch}{1.1}
\caption{Top-1 Accuracy of target model under different known label quantity}
\vskip -0.1in
\label{tab:1}
\centering
\setlength{\tabcolsep}{1.8mm}{
\begin{tabular}{c c c c c c c}
\toprule
\multirow{2}{*}{\textbf{\begin{tabular}[c]{@{}c@{}}Known Label \\ Quantity\end{tabular}}} & \multicolumn{2}{c}{\textbf{MNIST}} & \multicolumn{2}{c}{\textbf{CIFAR-10}} & \multicolumn{2}{c}{\textbf{CIFAR-100}} \\ \cmidrule(lr{0pt}){2-3} \cmidrule(lr{0pt}){4-5}  \cmidrule(lr{0pt}){6-7}
 & \textbf{Train} & \textbf{Test} & \textbf{Train} & \textbf{Test} & \textbf{Train} & \textbf{Test} \\ \midrule
10 & 0.5935 & 0.5105 & 0.6316 & 0.5373 & 0.5284 & 0.4253 \\
20 & 0.6382 & 0.5624 & 0.7325 & 0.5921 & 0.5495 & 0.4632 \\
40 & 0.6857 & 0.5759 & 0.7913 & 0.6192 & 0.6159 & 0.4967 \\
80 & 0.7205 & 0.5991 & 0.8305 & 0.6424 & 0.6793 & 0.5274 \\
160 & 0.7894 & 0.6312 & 0.8723 & 0.6814 & 0.7416 & 0.5934 \\
320 & \textbf{0.7953} & \textbf{0.6462} & \textbf{0.8836} & \textbf{0.6953} & \textbf{0.7528} & \textbf{0.6073} \\ \bottomrule
\end{tabular}}
\end{table}

Following the P-GAN approach detailed in section \ref{sec:P-GAN}, we poison 20\% of the local dataset during the training of VFL. As illustrated in Table \ref{tab:2}, our method substantially outperforms LRA and VILLAIN in reducing the model's $F_1$ score, showcasing its superior poisoning efficacy. Besides, DAE ensures that the top model retains a respectably good performance level when defending against poison attacks. DAE mitigates the effects of poisoning attacks, especially on the MNIST dataset, preserving a high classification level.

\begin{table}[ht]
\renewcommand{\arraystretch}{1.1}
\caption{$F_1$ scores of trained models with different methods.}
\vskip -0.1in
\label{tab:2}
\centering
\begin{tabular}{cccc}
\toprule
\textbf{Approach} & \textbf{MNIST} & \textbf{CIFAR-10} & \textbf{CIFAR-100} \\ \midrule
P-GAN & \textbf{0.8243} & \textbf{0.8091} & \textbf{0.5962} \\
LRA & 0.8522 & 0.8210 & 0.6083 \\
VILLAIN & 0.9029 & 0.8257 & 0.6294 \\
P-GAN+DAE & 0.9157 & 0.8883 & 0.6422 \\
LRA+DAE & 0.9245 & 0.8779 & 0.6684 \\
VILLAIN+DAE & 0.9326 & 0.8994 & 0.6728 \\
normal training & 0.9723 & 0.9451 & 0.7248 \\
 \bottomrule
\end{tabular}
\end{table}
\subsection{Sensitivity Evaluation}
In this section, we further investigate the factors that could influence the effectiveness of attack or defense. These factors cover hyperparameters, the number of adversary’s features, the quantity of known labels, and the poisoned sample ratio.

\begin{figure}[ht]
  \centering
  \subfigure[P-GAN]{
    \includegraphics[width=0.48\linewidth]{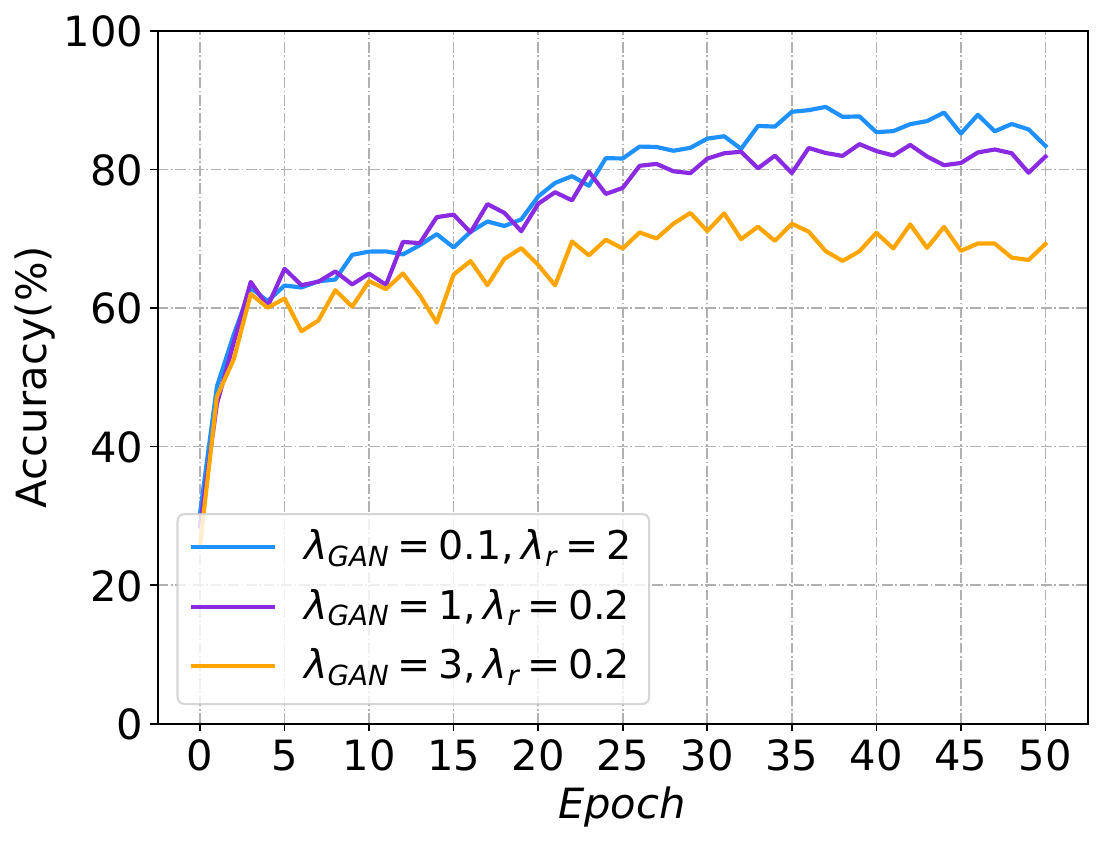}
    \label{fig:subfig1}
  }
  \hspace{-5mm} 
  \subfigure[PGAN+DAE]{
    \includegraphics[width=0.48\linewidth]{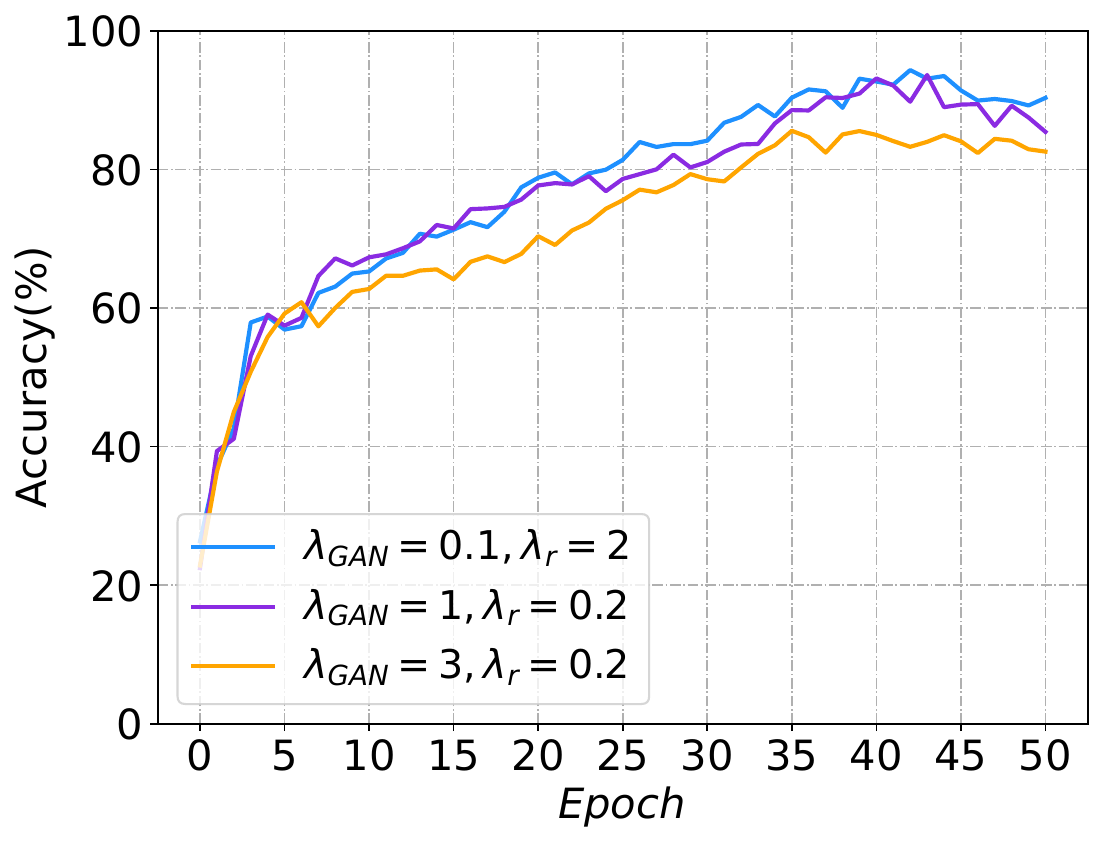}
    \label{fig:subfig2}
  }
  \caption{Test accuracy of the VFL model on MNIST under various parameter settings when subjected to P-GAN poisoning attacks and defended by DAE.}
  \label{fig:mainfig4}
\end{figure}

\textbf{Impact of hyperparameters.} We evaluate the performance of P-GAN and DAE on MNIST under various settings of $\lambda_{r}$ and $\lambda_{GAN}$. As illustrated in Figure \ref{fig:mainfig4}, the choice of these parameters can impact the loss optimization items' priorities. Notably, with an increased $\lambda_{GAN}$, the poisoning model tends to produce samples with high attack success, degrading the top model performance. Due to the more significant perturbations, DAE shows enhanced effectiveness in mitigating the poisoning effects. Conversely, with an increased number of $\lambda_{r}$,  the poisoning model favors samples with better disguise, resulting in better top model performance and greater resistance to DAE.

\begin{figure}[ht]
  \centering
  \subfigure[MNIST]{
    \includegraphics[width=0.48\linewidth]{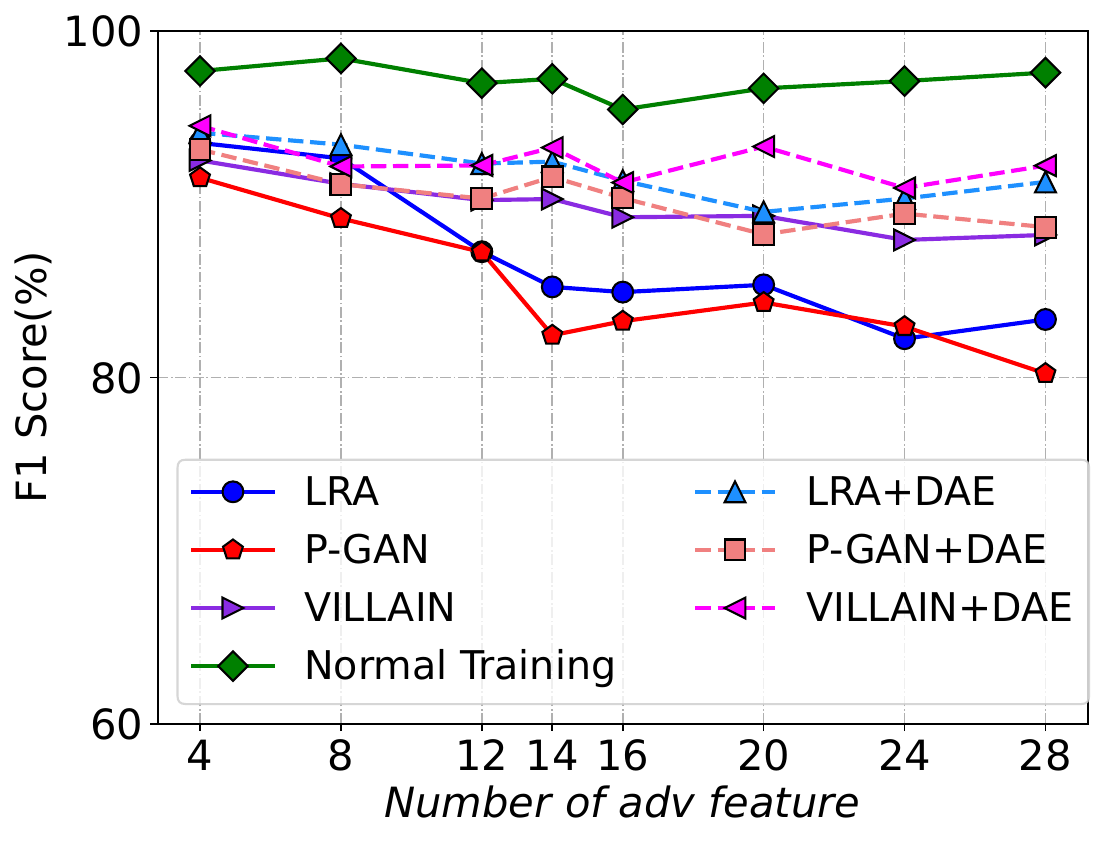}
    \label{fig:subfig3-1}
  }
  \hspace{-5mm} 
  \subfigure[CIFAR-10]{
    \includegraphics[width=0.48\linewidth]{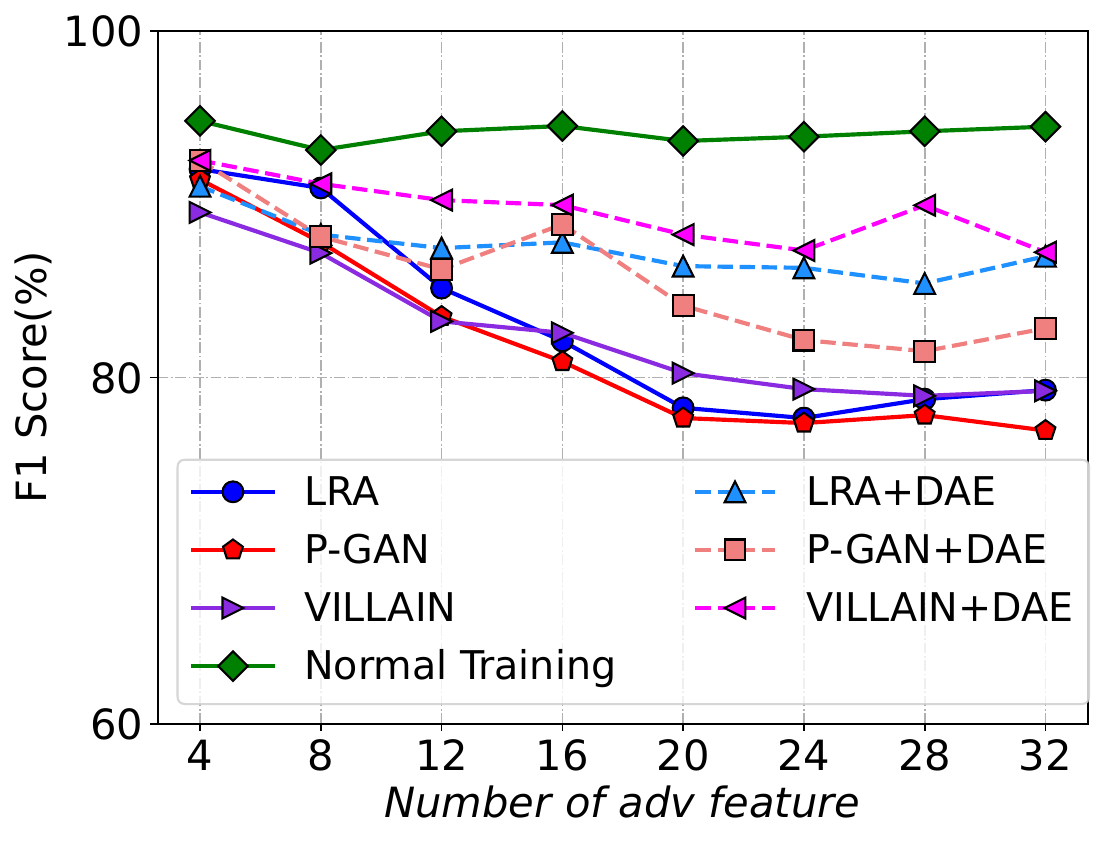}
    \label{fig:subfig3-2}
  }
  \caption{$F_1$ scores of poisoned model under varied number of adversary's features, with a height pixel interval of 4.}
  \label{fig:mainfig1}
\end{figure}

\textbf{Number of the adversary's features.} In VFL, the malicious participant aims to degrade the performance of the top model with its local features. We evaluate model performance under different number of adversary's features. Experiments are conducted on the MNIST and CIFAR-10, with segmenting images based on feature quantity. For example, on CIFAR-10, the malicious participant holds image slices varying between 4 and 28 pixels in height. As illustrated in Figure \ref{fig:mainfig1}, the results indicate that as the number of adversary's features grows, the performance of the top model declines further. When the malicious participant controls over 50\% of the features, the top model's $F_1$ score consistently remains at a relatively low level. Besides, compared to LRA, P-GAN exhibits enhanced attack efficacy, resulting in a more significant decrease in the model's F1 score. This occurs as P-GAN effectively captures the intrinsic representations of local data to generate poisoned samples, leading to reduced feature reconstruction errors. Moreover, as the number of features held by the malicious participant increases, the efficacy of the poisoning attack improves due to the malicious entities being able to perturb a wider range of critical features. The DAE method consistently mitigates the effects of poisoning attacks under various scenarios.

\begin{figure}[ht]
  \centering
  \subfigure[MNIST]{
    \includegraphics[width=0.48\linewidth]{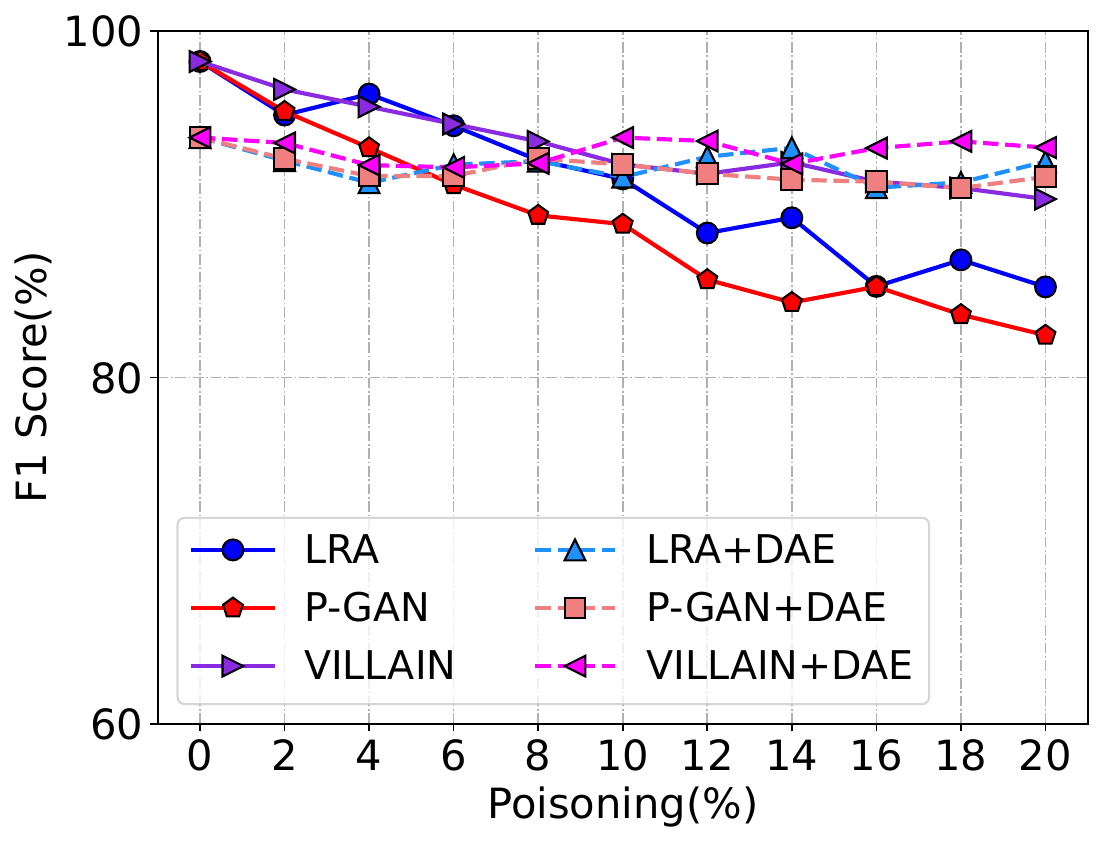}
    \label{fig:subfig2-1}
  }
  \hspace{-5mm} 
  \subfigure[CIFAR-10]{
    \includegraphics[width=0.48\linewidth]{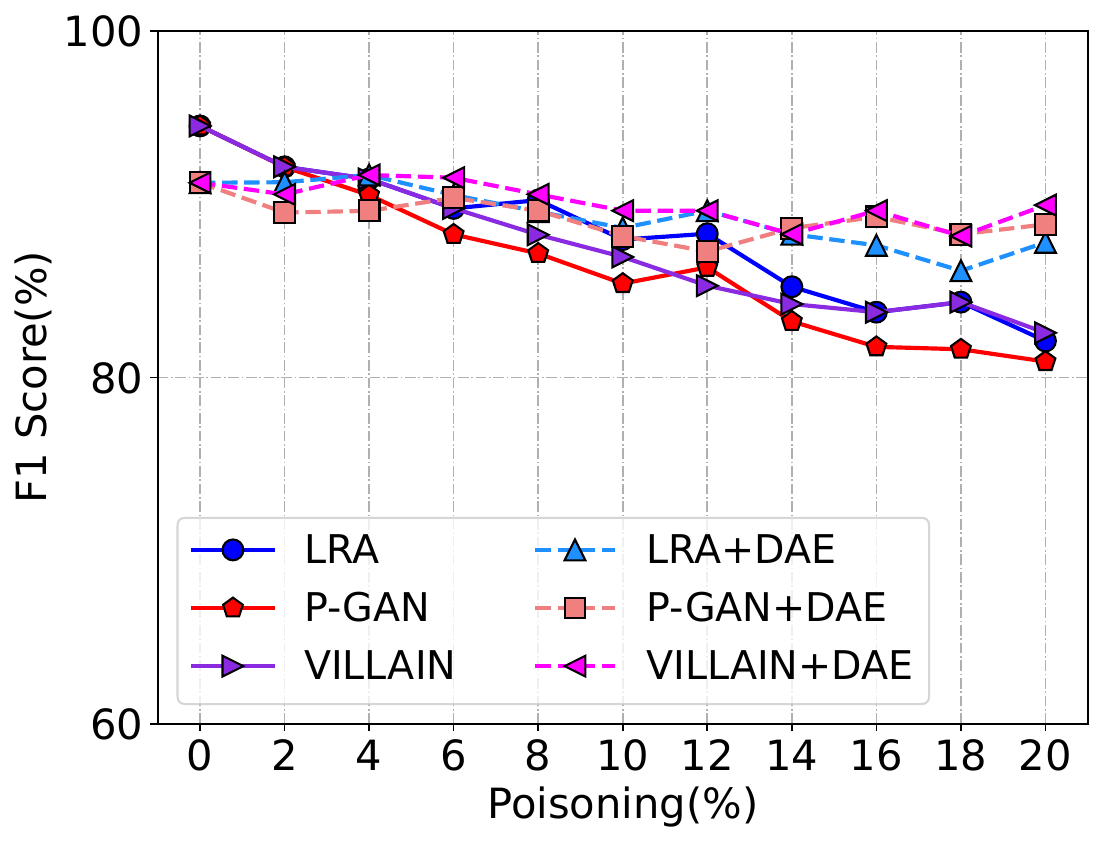}
    \label{fig:subfig2-2}
  }
  \caption{$F_1$ Scores of poisined models with varying the percentage of poisoning data from 0\% to 20\% with step 2\%.}
  \label{fig:mainfig2}
\end{figure}

\textbf{Percentage of poisoned data} 
We conduct a comparative analysis of the percentage of poisoned data. The accuracy of the poisoned models across poisoning proportions ranging from 0\% to 20\% is detailed in Figure \ref{fig:mainfig2}. As the proportion of poisoned samples increases, the top model's performance notably deteriorates, especially when subjected to P-GAN. DAE consistently alleviates the adversarial impact of LRA and P-GAN. However, upon employing the DAE method for server-side outlier detection, the performance of the top model shows a slow decline as the proportion of poisoned samples increases. This indicates that DAE's resistance to poisoning attacks diminishes as the poisoned data in the training set grows.

\begin{figure}[ht]
  \centering
  \subfigure[MNIST]{
    \includegraphics[width=0.48\linewidth]{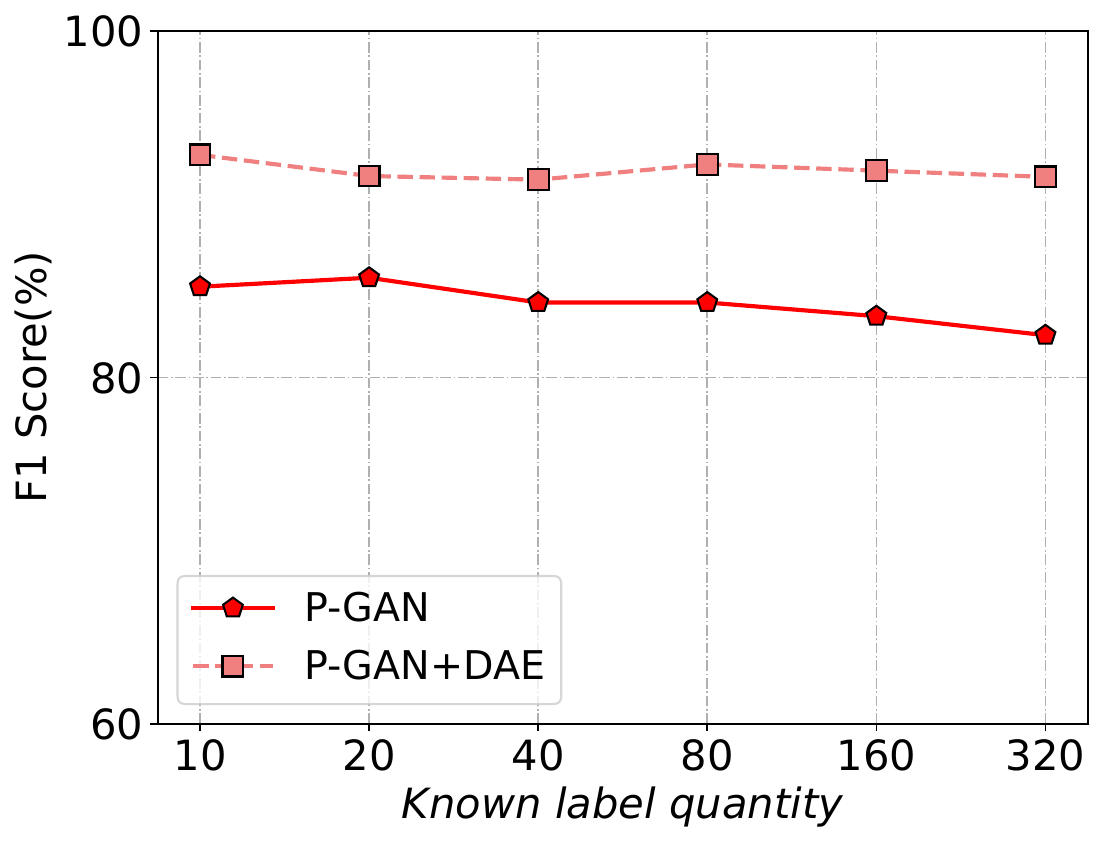}
    \label{fig:subfig1-11}
  }
  \hspace{-5mm} 
  \subfigure[CIFAR-10]{
    \includegraphics[width=0.48\linewidth]{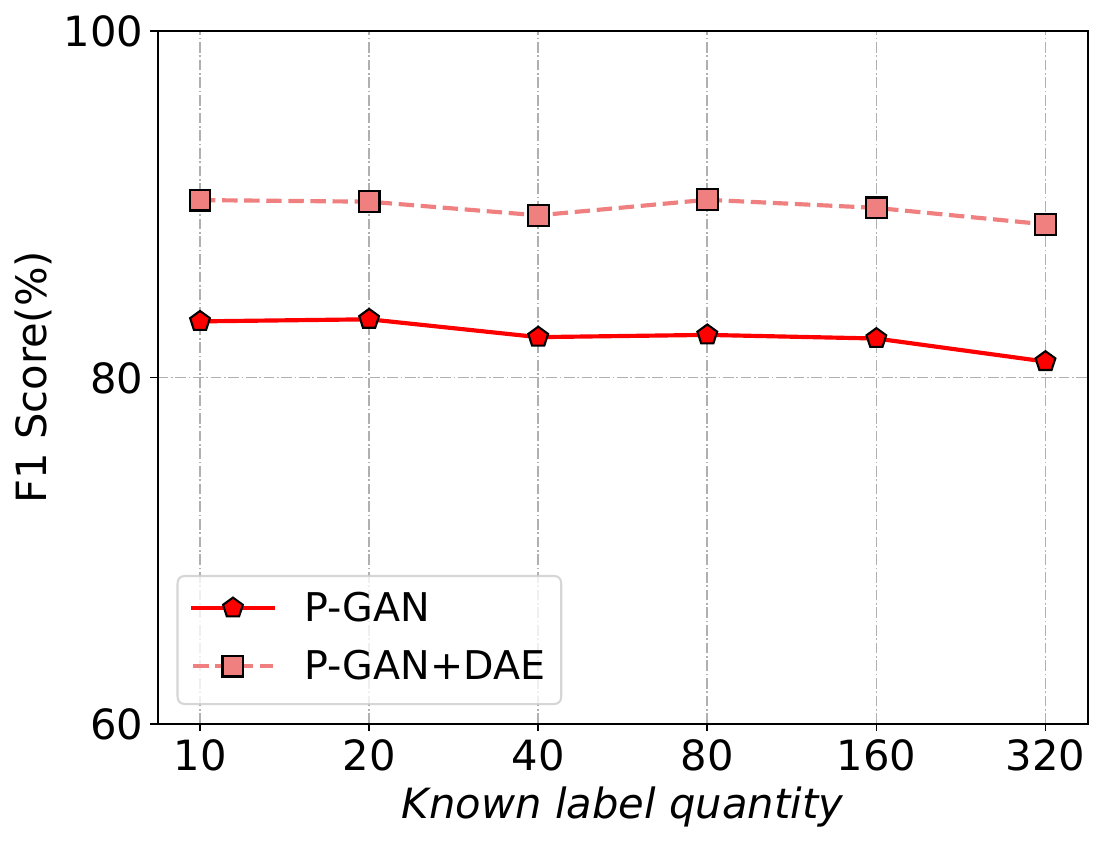}
    \label{fig:subfig1-2}
  }
  \caption{$F_1$ scores of the poisoned models with varying amounts of labeled samples known locally by adversary.}
  \label{fig:mainfig3}
\end{figure}

\textbf{Quantity of known labels.} We evaluate the influence of known labels available to the malicious participant on the target model. With varying amounts of known labels, we obtain the surrogate target models by Fixmatch, and the performance of these poisoned top models is presented in Figure \ref{fig:mainfig3}. As the quantity of known labels increases, P-GAN further diminishes poisoned model performance. In terms of defense, the DAE maintains a high model performance by detecting outlier data. As the number of known labels increases, its defensive performance gradually diminishes. This indicates that, with increased known labels, P-GAN becomes less effective against DAE defense within a certain range.
\section{Conclusion}
In this paper, we present P-GAN, an end-to-end poisoning attack framework tailored for vertical federated learning. Utilizing the Fixmatch method, we obtain a surrogate target model for the poisoning attack. Subsequently, we employ a GAN-based perturbation model to produce poisoned samples. During the federated learning training process, these poisoned samples degrade the performance of the top model.
On the server side, we introduce a DAE-based anomaly detection framework tailored for vertical federated learning. This framework uses reconstruction errors of the embedding vectors and label information to filter out outlier data. 
We conduct comprehensive experiments to validate the effectiveness of the P-GAN and DAE. Considering potential malicious participants during the training phase, the server should utilize effective methods like DAE to defend against the poisoning threats from participants.

\section*{Acknowledgment}
This research was supported by the National Key Research and Development Program of China, under Grant No. 2022ZD0120201 - ``Unified Representation and Knowledge Graph Construction for Science Popularization Resources''.

\bibliographystyle{IEEEtran}
\bibliography{IEEEexample}

\begin{thebibliography}{10}
\providecommand{\url}[1]{#1}
\csname url@samestyle\endcsname
\providecommand{\newblock}{\relax}
\providecommand{\bibinfo}[2]{#2}
\providecommand{\BIBentrySTDinterwordspacing}{\spaceskip=0pt\relax}
\providecommand{\BIBentryALTinterwordstretchfactor}{4}
\providecommand{\BIBentryALTinterwordspacing}{\spaceskip=\fontdimen2\font plus
\BIBentryALTinterwordstretchfactor\fontdimen3\font minus \fontdimen4\font\relax}
\providecommand{\BIBforeignlanguage}[2]{{%
\expandafter\ifx\csname l@#1\endcsname\relax
\typeout{** WARNING: IEEEtran.bst: No hyphenation pattern has been}%
\typeout{** loaded for the language `#1'. Using the pattern for}%
\typeout{** the default language instead.}%
\else
\language=\csname l@#1\endcsname
\fi
#2}}
\providecommand{\BIBdecl}{\relax}
\BIBdecl

\bibitem{EU}
P.~Regulation, ``Regulation (eu) 2016/679 of the european parliament and of the council,'' \emph{Regulation (eu)}, vol. 679, p. 2016, 2016.

\bibitem{mcmahan2017communication}
B.~McMahan, E.~Moore, D.~Ramage, S.~Hampson, and B.~A. y~Arcas, ``Communication-efficient learning of deep networks from decentralized data,'' in \emph{Artificial intelligence and statistics}.\hskip 1em plus 0.5em minus 0.4em\relax PMLR, 2017, pp. 1273--1282.

\bibitem{niknam2020federated}
S.~Niknam, H.~S. Dhillon, and J.~H. Reed, ``Federated learning for wireless communications: Motivation, opportunities, and challenges,'' \emph{IEEE Communications Magazine}, vol.~58, no.~6, pp. 46--51, 2020.

\bibitem{konevcny2016federated}
J.~Kone{\v{c}}n{\`y}, H.~B. McMahan, F.~X. Yu, P.~Richt{\'a}rik, A.~T. Suresh, and D.~Bacon, ``Federated learning: Strategies for improving communication efficiency,'' \emph{arXiv preprint arXiv:1610.05492}, 2016.

\bibitem{kairouz2021advances}
P.~Kairouz, H.~B. McMahan, B.~Avent, A.~Bellet, M.~Bennis, A.~N. Bhagoji, K.~Bonawitz, Z.~Charles, G.~Cormode, R.~Cummings \emph{et~al.}, ``Advances and open problems in federated learning,'' \emph{Foundations and Trends{\textregistered} in Machine Learning}, vol.~14, no. 1--2, pp. 1--210, 2021.

\bibitem{yang2019federated}
Q.~Yang, Y.~Liu, T.~Chen, and Y.~Tong, ``Federated machine learning: Concept and applications,'' \emph{ACM Transactions on Intelligent Systems and Technology (TIST)}, vol.~10, no.~2, pp. 1--19, 2019.

\bibitem{cheng2021secureboost}
K.~Cheng, T.~Fan, Y.~Jin, Y.~Liu, T.~Chen, D.~Papadopoulos, and Q.~Yang, ``Secureboost: A lossless federated learning framework,'' \emph{IEEE Intelligent Systems}, vol.~36, no.~6, pp. 87--98, 2021.

\bibitem{wei2021autoheri}
P.~Wei, W.~Zhang, Z.~Xu, S.~Liu, K.-c. Lee, and B.~Zheng, ``Autoheri: Automated hierarchical representation integration for post-click conversion rate estimation,'' in \emph{Proceedings of the 30th ACM International Conference on Information \& Knowledge Management}, 2021, pp. 3528--3532.

\bibitem{zhu2019deep}
L.~Zhu, Z.~Liu, and S.~Han, ``Deep leakage from gradients,'' \emph{Advances in neural information processing systems}, vol.~32, 2019.

\bibitem{weng2020privacy}
H.~Weng, J.~Zhang, F.~Xue, T.~Wei, S.~Ji, and Z.~Zong, ``Privacy leakage of real-world vertical federated learning,'' \emph{arXiv preprint arXiv:2011.09290}, 2020.

\bibitem{fu2022label}
C.~Fu, X.~Zhang, S.~Ji, J.~Chen, J.~Wu, S.~Guo, J.~Zhou, A.~X. Liu, and T.~Wang, ``Label inference attacks against vertical federated learning,'' in \emph{31st USENIX Security Symposium (USENIX Security 22)}, 2022, pp. 1397--1414.

\bibitem{luo2021feature}
X.~Luo, Y.~Wu, X.~Xiao, and B.~C. Ooi, ``Feature inference attack on model predictions in vertical federated learning,'' in \emph{2021 IEEE 37th International Conference on Data Engineering (ICDE)}.\hskip 1em plus 0.5em minus 0.4em\relax IEEE, 2021, pp. 181--192.

\bibitem{yang2023practical}
R.~Yang, J.~Ma, J.~Zhang, S.~Kumari, S.~Kumar, and J.~J. Rodrigues, ``Practical feature inference attack in vertical federated learning during prediction in artificial internet of things,'' \emph{IEEE Internet of Things Journal}, 2023.

\bibitem{liu2021batch}
Y.~Liu, T.~Zou, Y.~Kang, W.~Liu, Y.~He, Z.~Yi, and Q.~Yang, ``Batch label inference and replacement attacks in black-boxed vertical federated learning,'' \emph{arXiv preprint arXiv:2112.05409}, 2021.

\bibitem{bai2023villain}
Y.~Bai, Y.~Chen, H.~Zhang, W.~Xu, H.~Weng, and D.~Goodman, ``$\{$VILLAIN$\}$: Backdoor attacks against vertical split learning,'' in \emph{32nd USENIX Security Symposium (USENIX Security 23)}, 2023, pp. 2743--2760.

\bibitem{yu2023backdoor}
F.~Yu, L.~Wang, B.~Zeng, K.~Zhao, Z.~Pang, and T.~Wu, ``How to backdoor split learning.'' \emph{Neural Networks: the Official Journal of the International Neural Network Society}, vol. 168, pp. 326--336, 2023.

\bibitem{sohn2020fixmatch}
K.~Sohn, D.~Berthelot, N.~Carlini, Z.~Zhang, H.~Zhang, C.~A. Raffel, E.~D. Cubuk, A.~Kurakin, and C.-L. Li, ``Fixmatch: Simplifying semi-supervised learning with consistency and confidence,'' \emph{Advances in neural information processing systems}, vol.~33, pp. 596--608, 2020.

\bibitem{liu2016delving}
Y.~Liu, X.~Chen, C.~Liu, and D.~Song, ``Delving into transferable adversarial examples and black-box attacks,'' \emph{arXiv preprint arXiv:1611.02770}, 2016.

\bibitem{bovenzi2022data}
G.~Bovenzi, A.~Foggia, S.~Santella, A.~Testa, V.~Persico, and A.~Pescap{\'e}, ``Data poisoning attacks against autoencoder-based anomaly detection models: a robustness analysis,'' in \emph{ICC 2022-IEEE International Conference on Communications}.\hskip 1em plus 0.5em minus 0.4em\relax IEEE, 2022, pp. 5427--5432.

\end{thebibliography}

\end{document}